\pdfoutput=1

\documentclass[11pt]{article}

\usepackage{EACL2023}

\usepackage{times}
\usepackage{latexsym}

\usepackage[T1]{fontenc}

\usepackage[utf8]{inputenc}

\usepackage{microtype}
\usepackage{xspace}
\usepackage{inconsolata}
\usepackage{tabularx}

\usepackage{amsmath}
\usepackage{graphicx}
\usepackage{tabularx}
\usepackage{xspace}
\usepackage[obeyfinal]{todo}
\usepackage{stfloats}
\usepackage[group-separator={,},group-minimum-digits={3}]{siunitx}

\newcommand{\src}{\ensuremath{\mathbf{f}}} 
\newcommand{\trg}{\ensuremath{\mathbf{e}}} 
\newcommand{\simsrc}{\ensuremath{\mathbf{\tilde{f}}}} 
\newcommand{\simtrg}{\ensuremath{\mathbf{\tilde{e}}}} 
\newcommand{\initrg}{\ensuremath{\mathbf{e^{\prime}}}} 
\newcommand{\initrgp}{\ensuremath{\mathbf{e^{\prime\prime}}}} 

\newcommand{\ourmodel}{TM-LevT\xspace}

\title{Integrating Translation Memories into Non-Autoregressive Machine Translation}

\author{Jitao Xu$^{\dag}$\ $\quad\quad\quad\quad$ Josep Crego$^{\ddag}$\ $\quad\quad\quad\quad$  François Yvon$^{\dag}$ \\ \\
  $^\dag$Universit\'e Paris-Saclay, CNRS, LISN, 91400, Orsay, France \\
  $^\ddag$SYSTRAN, 5 rue Feydeau, 75002, Paris, France \\
  \texttt{\{jitao.xu,francois.yvon\}@limsi.fr, josep.crego@systrangroup.com} \\}

\begin{document}
\maketitle
\begin{abstract}
  Non-autoregressive machine translation (NAT) has recently made great progress. However, most works to date have focused on standard translation tasks, even though some edit-based NAT models, such as the Levenshtein Transformer (LevT), seem well suited to translate with a Translation Memory (TM). This is the scenario considered here. We first analyze the vanilla LevT model and explain why it does not do well in this setting. We then propose a new variant, TM-LevT, and show how to effectively train this model. By modifying the data presentation and introducing an extra deletion operation, we obtain performance that are on par with an autoregressive approach, while reducing the decoding load. We also show that incorporating TMs during training dispenses to use knowledge distillation, a well-known trick used to mitigate the multimodality issue.
\end{abstract}

\section{Introduction\label{sec:intro}}

Non-autoregressive neural machine translation (NAT) has been greatly advanced in recent years \citep{Xiao22surveyNAT}. NAT takes advantage from parallel decoding to generate multiple tokens simultaneously and speed up inference. This is often at the cost of a loss in translation quality when compared to autoregressive (AR) models \citep{Gu18nonautoregressive}. This gap is slowly closing and methods based on iterative refinement \citep{Ghazvininejad19mask,Gu19levenshtein,Saharia20nonauto} and on connectionist temporal classification \citep{Libovicky18end,Gu21fully} are now reporting BLEU scores similar to strong AR baselines.

Most works on NAT focus on the standard machine translation (MT) task, where the decoder starts from scratch, with the exception of \citet{Susanto20lexically,Xu21editor}, who use NAT to integrate lexical constraints in decoding. However, edit-based NAT models, such as the Levenshtein Transformer (LevT) of \citet{Gu19levenshtein}, seem to be a natural candidate to perform MT with Translation Memories (TM). LevT is able to iteratively edit an initial target sequence by performing insertion and deletion operations until convergence. This design also matches the concept of using TMs in MT, where given a source sentence, we aim to edit a candidate translation retrieved from the TM.

This idea has been used for decades in the localization industry and implemented into basic Computer-Aided Translation tools. Translators wishing to translate a sentence can benefit from fuzzy matching techniques to retrieve similar segments from the TM. These segments can then be revised, thereby improving productivity and consistency of the translation process \citep{Koehn10convergence,Yamada11effect}. The retrieval of similar examples from a TM has also proved useful in conventional (AR) neural MT systems; they can be injected into the encoder \citep{Bulte19fuzzy,Xu20boosting} or as priming signals in the decoder \citep{Pham20priming} to influence the translation process. These studies report significant gains in translation performance in technical domains, where the translation of terms and phraseology greatly benefits from examples found in a TM.

Our main focus in this work is to develop an improved version of LevT suited to the \emph{revision part} of TM use, where the translation retrieved from TM is modified via edit operations in a non-autoregressive way. We first show that the original LevT cannot perform well on this task and explain that this failure is a direct consequence of its training design. We propose to fix this issue with \ourmodel, which includes an additional deletion step. Next, we propose to further improve the training procedure in two ways: (a) by also including the retrieved candidate translation on the source side, as done in AR TM-based approaches \citep{Bulte19fuzzy,Xu20boosting}; (b) by simultaneously training with empty and non-empty initial target sentences. In our experiments, \ourmodel{} achieves performance that is on par with a strong AR approach on various domains when translating with TMs, with a reduced decoding load. We also observe that incorporating an initial translation both on the source and target sides makes Knowledge Distillation (KD, \citealp{Kim16sequence}) useless. This contrasts with standard NAT models, which rely on KD to alleviate the multimodality issue \citep{Gu18nonautoregressive}. As far as we know, this work is the first to study NAT with TMs in a controlled setting.

Our contributions are the following: (a) we show that the original LevT training scheme is not suited to edit similar translations from a TM; (b) we propose a variant of LevT, \ourmodel{} with an improved training procedure, which yields performance that are close, or even similar to AR approaches when translating with good TM matches, with a reduced decoding load; (c) we highlight the benefits of multi-task training (with and without TMs) to attain the best performance; (d) we discuss the reasons why KD hurts the training of NAT with TMs.

\section{Using Translation Memories in NAT\label{sec:method}}

\subsection{Background\label{ssec:background}}

\paragraph{TM Retrieval}

Given a source sentence \src, we aim to retrieve a good match \simtrg{} from the TM. For this, we search the TM for a pair of sentences $(\simsrc, \simtrg)$, where \simsrc{} is similar to \src. The corresponding target \simtrg{} is then used to initiate the translation of \src. We compute the similarity between \src{} and \simsrc{} as:
\begin{equation}
  \operatorname{sim}(\src, \simsrc) = 1 - \frac{\operatorname{ED}(\src, \simsrc)}{\max(|\src|, |\simsrc|)},\label{eq:similarity}
\end{equation}
where $\operatorname{ED}(\src, \simsrc)$ is the edit distance between \src{} and \simsrc, and $|\src|$ is the length of \src. The intuition is that the closer \src{} and \simsrc{} are, the more suitable \simtrg{} will be. As is custom, we only use TM matches when the similarity score exceeds a predefined threshold, otherwise we translate from scratch. We discuss the effect of the match similarity in Section~\ref{ssec:threshold}.

\paragraph{Levenshtein Transformer}

LevT is an edit-based NAT model proposed by \citet{Gu19levenshtein}. It performs translation by iteratively editing an initial target sequence with insertion and deletion operations until convergence. The insertion operation is composed of a \emph{placeholder insertion module} and a \emph{token predictor}. The placeholder classifier predicts the number of additional tokens that need to be inserted between any two consecutive tokens in its input sequence. The token predictor then generates a token for each placeholder position. The deletion operation aims to detect prediction errors made by the model. It makes a binary decision for each token, indicating whether it should be deleted or kept. During training, a noised initial target sequence \initrg{} is first generated by randomly dropping tokens from the reference \trg. The insertion modules learn to reinsert the deleted tokens into \initrg. The deletion operation is then trained to erase erroneous predictions made during insertion.

During inference, LevT starts with an empty target sequence ($\initrg = []$) and generates the translation by alternatively performing deletion and insertion operations until convergence or a maximum number of decoding rounds is reached. In the first iteration, the deletion is omitted, as no tokens can be deleted from the empty sequence. This iterative refinement procedure converges when the input and output of one iteration are the same, either because LevT predicts nothing to delete and to insert, or because it enters a loop where the deleted tokens are reinserted in the same round. Unlike almost all other NAT models, LevT does not require any external prediction of the target length, as the number of target tokens is iteratively revised and adjusted by the placeholder prediction module. We refer to \citet{Gu19levenshtein} for more details about LevT.

\begin{figure*}[ht]
  \center
  \includegraphics[width=0.92\textwidth]{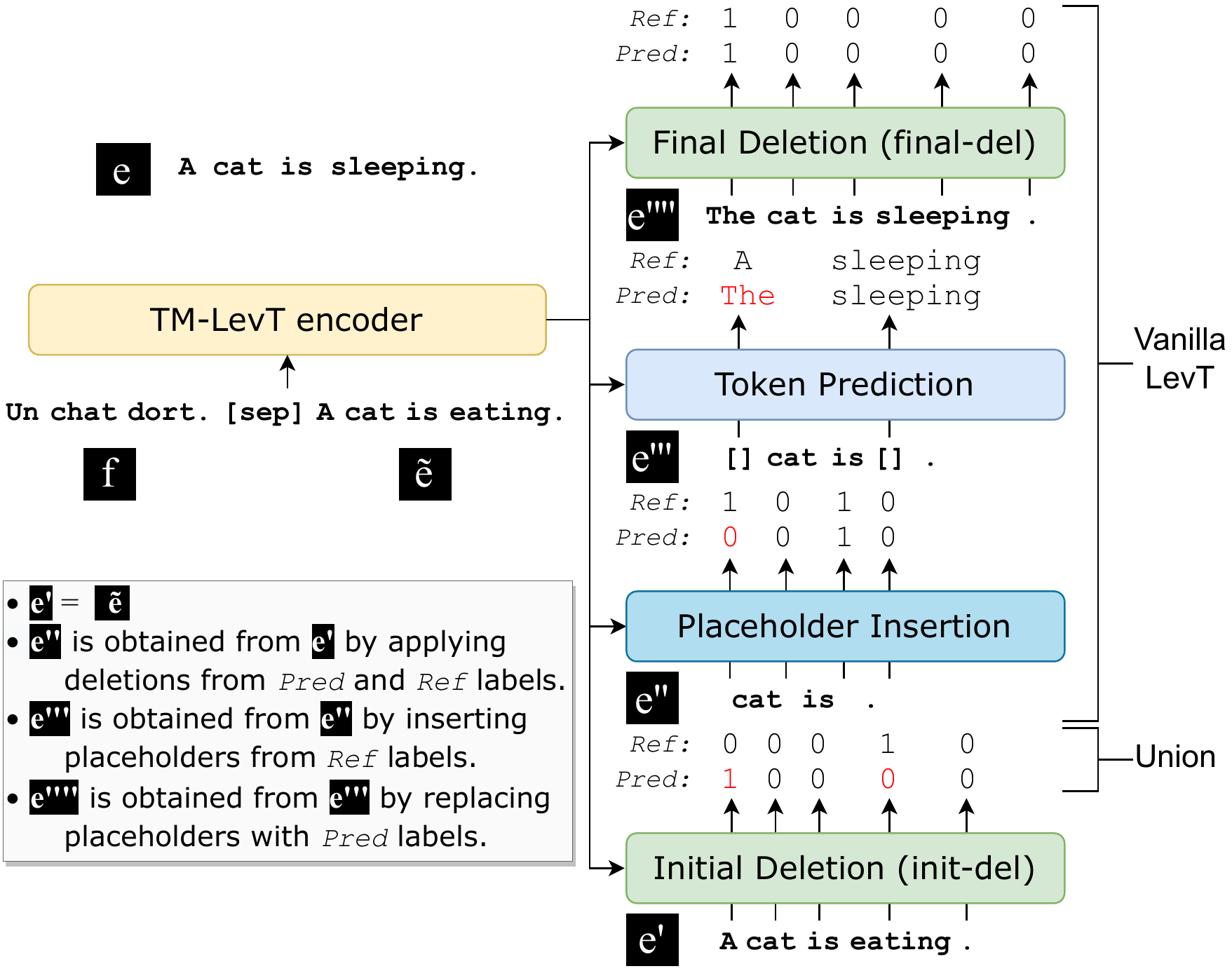}
  \caption{A complete training step for \ourmodel. Compared to the original LevT which starts training from \initrgp, \ourmodel{} adds the init-del step to delete unrelated tokens from a TM match. Figure better viewed in color.\label{fig:tm-levt}}
\end{figure*}

\subsection{Deficiencies of LevT Training\label{ssec:fluency}}

Even though the edit-based nature of LevT makes it readily able to translate with TMs, it has mostly been applied to standard MT, where the decoder starts with an empty sentence.\footnote{One notable exception is the attempt in \citet{Gu19levenshtein} to perform automatic post-editing through iterative revisions.} This is consistent with the overall training scheme, illustrated in Figure~\ref{fig:tm-levt} (Vanilla LevT), where inputs for the placeholder insertion module are always subsequences of the reference and the deletion module only sees the outputs of the previous token insertion step.

\begin{table}[!ht]
  \center
  \scalebox{0.85}{
  \begin{tabular}{l|ccc}
  \hline
  Settings & Empty & Random Sent & Shuffle Ref \\
  \hline
  Init & - & 1.3 & 5.0 \\
  LevT & 45.4 & 2.1 & 40.2 \\
  \hline
  LevT vs Init & - & 90.4 & 9.4 \\
  \hline
  \end{tabular}
  }
  \caption{BLEU scores of LevT decoding with various target initialization. \textit{Empty} refers to standard LevT inference with an empty start. \textit{Random Sent} uses a random sentence as initial target. \textit{Shuffle Ref} starts with a random shuffle of the reference translation. \textit{Init} reports the BLEU score of the initialization, while \textit{LevT vs Init} compares LevT's outputs with their starting points.\label{tab:fluency}}
\end{table}

To illustrate the deficiency of this training scheme, we learned a vanilla LevT model using the datasets of Section~\ref{ssec:data} and initialized the decoder with a sentence randomly selected from the test set and totally unrelated to the source sentence. We observe (Table~\ref{tab:fluency}, Random Sent) that LevT's outputs are almost as bad as their starting point. This is because the deletion module fails to delete irrelevant input words, presenting the insertion modules with a fully fluent yet fully inadequate sequence that the insertion module is hard-pressed to revise. This contrasts with the \textit{Shuffle Ref} scenario, where the decoder starts with a random shuffle of the reference. LevT can now make changes during the iterative refinement and generates translations ($40.2$ BLEU) that are close to standard decoding ($45.4$ BLEU). The TM-based scenario discussed below presents the same challenge for the deletion module, that of spotting and deleting irrelevant words. Our proposal will first focus on fixing this issue.

\subsection{Improving Editions with \ourmodel\label{ssec:tmlevt}}

The experiment of previous section suggests that LevT will have issues editing TM matches, as they often contain tokens that are unrelated to the source and should be removed (see Figure~\ref{fig:tm-levt} for an example TM match \simtrg{} containing an unrelated word "eating"). The distribution of unrelated tokens may greatly differ from token prediction errors made by LevT, which are tokens LevT is trained to delete.

We propose a variant of LevT denoted \ourmodel{}, where we include an extra deletion step (init-del) that applies before the insertion modules. As shown in Figure~\ref{fig:tm-levt}, init-del is trained to detect unrelated tokens from the initial \initrg, whereas the final deletion (final-del) focuses on prediction errors. During training, we generate examples for the insertion modules by removing from \initrg{} tokens that either are not in the reference, or should be deleted according to the init-del operation. The resulting subsequence $\initrgp$ is then used to train the insertion operation. \ourmodel{} does not change the number of parameters, as we use the same classifier for the init-del and final-del steps. During inference, \ourmodel{} behaves exactly as LevT, iteratively applying deletions and insertions to an initial candidate translation.

\subsection{Translating with or without TM Matches}

In practical applications, two modes of operations need to be supported. The first is when a good match is found in the TM and used to initialize the decoding $(\initrg = \simtrg)$. Revising $\initrg$ may imply to delete or insert tokens, which is what the system is trained for. It may also imply to move words around, which is achieved by a succession of deletion and insertion. However, as these operations are performed independently, there is no guarantee that the deleted words will be memorized for subsequent insertions, causing the loss of relevant words in the process. This is again illustrated in Table~\ref{tab:fluency} (Shuffle Ref) where we see that even when given all the reference tokens (in random order), LevT still underperforms translating from scratch. To mitigate this risk, we augment the source side with the initial TM match, ensuring that \simtrg{} is always fully available to the decoder. Following the proposal of \citet{Bulte19fuzzy,Xu20boosting} for AR models, this is performed by concatenating \src{} and \simtrg{} on the encoder side (see Figure~\ref{fig:tm-levt}).

The second mode of operation is when no appropriate match is found, causing the system to fall back to a standard MT regime. In order to handle both situations in a single model, we resort to multi-task training and prepare our training samples as follows: with probability $p=0.5$, we decide either to decode with a retrieved TM match \simtrg{} or from scratch. In the former case, the decoder is initialized with \simtrg, while in the latter, we use a noised subsequence \initrg{} generated as in \citep{Gu19levenshtein}. \ourmodel{} is then jointly trained on both tasks.

\section{Experiments\label{sec:exp}}

\subsection{Datasets\label{ssec:data}}

Our experiments use the same corpus as \citet{Xu20boosting}, and contains texts from a diverse set of $11$ domains for the English-French direction, downloaded from OPUS\footnote{\url{https://opus.nlpl.eu/}} \citep{Tiedemann12parallel}: documents from the European Central Bank (ECB); from the European Medicines Agency (EMEA); Proceedings of the European Parliament (Epps); legislative texts of the European Union (JRC); News Commentaries (News); TED talk subtitles (TED); parallel sentences extracted from Wikipedia (Wiki); localization files (GNOME, KDE and Ubuntu) and manuals (PHP). We include both technical domains, for which good matches are likely to exist, and more "general" domains (Epps and News), for which useful TM matches are harder to find, allowing us to explore the benefits of using TMs for a variety of conditions. All these data are deduplicated prior to training. 

For each source sentence, we retrieve from the same domain the top $3$ TM matches according to the similarity score of Equation~\eqref{eq:similarity} based on Fuzzy-Match toolkit\footnote{\url{https://github.com/SYSTRAN/fuzzy-match}}, further requiring a score of $0.4\leq\operatorname{sim}$$<$$1$. For each domain, we prepare two test sets with \num{1000} sentences each: one contains randomly selected sentences with a close match ($\operatorname{sim}$$>$$0.6$) in the TM, the other with an acceptable match ($\operatorname{sim}$$\in$$[0.4, 0.6]$). We also leave a held-out set of \num{1000} sentences per domain, for which no matches of $\operatorname{sim}$$\geq$$0.4$ are found. The remaining data is used for training. Note that the ratio of sentences with at least one TM match greatly varies across domains. Detailed statistics about these corpora are in Appendix~\ref{sec:data}. We use all retrieved TM matches (up to $3$) for training and only the best match for test. Therefore, a source training sentence with $3$ TM matches yields $3$ training samples. The initial set of $4.4$M parallel data is thus extended with about $5$M examples augmented with a TM match. We tokenize all data using the Moses tokenizer\footnote{\url{https://github.com/moses-smt/mosesdecoder}} and build a shared source-target vocabulary with $32$K Byte Pair Encoding units \citep{Sennrich16BPE} learned with \texttt{subword-nmt}.\footnote{\url{https://github.com/rsennrich/subword-nmt}}

\subsection{Experimental Settings\label{ssec:exp}}

We compare \ourmodel{} with a strong AR approach \citep{Bulte19fuzzy} and the original LevT model.\footnote{\url{https://github.com/facebookresearch/fairseq/tree/main/examples/nonautoregressive_translation}} These baselines use the same training data as \ourmodel, and also process examples with and without TM matches. For the AR model, TM matches only appear concatenated to the source sentence and translation always starts from scratch; for LevT, we test both cases where the decoder is initialized with and without TM matches.

\ourmodel{} is based on the Transformer architecture \citep{Vaswani17attention}, implemented with fairseq\footnote{\url{https://github.com/pytorch/fairseq}} \citep{Ott19fairseq}.\footnote{Our implementation is open-sourced at \url{https://github.com/jitao-xu/tm-levt}.} We use a hidden size of \num{512} and a feedforward size of \num{2048}, optimizing with Adam with a maximum learning rate of $0.0005$, an inverse square root decay schedule, and \num{10000} warm-up steps. We share the decoder parameters for both two deletions and the insertion operation and also tie all input and output embedding matrices \citep{Press17using}. We train \ourmodel{} for \num{300}k iterations with a batch size of \num{8192} tokens per GPU on $8$ V100 GPUs. For inference, we use a maximum iteration number of $10$ for \ourmodel and LevT, and a beam size of $5$ for the AR decoder. We use a batch size of \num{8192} tokens and perform inference on one single GPU for all compared models. The vanilla LevT model is trained similarly, while the AR model \citep{Bulte19fuzzy} is trained with a maximum learning rate of $0.0007$, with \num{4000} warmup steps for \num{300}k iterations on $4$ V100 GPUs. We report results of a \textit{do-nothing} baseline which simply copies the TM matches as outputs. Performance is measured with BLEU using SacreBLEU\footnote{SacreBLEU signature: BLEU+case.mixed+lang.en-fr+numrefs.1+smooth.exp+tok.13a+version.1.5.1} \citep{Post18sacrebleu} and with COMET \citep{Rei20comet}.

\section{Results and Analyses\label{sec:results}}

\subsection{NAT Can Benefit from TMs\label{ssec:results}}

\begin{table*}[!ht]
  \center
  \scalebox{0.85}{
  \begin{tabular}{l|ccccccccccc|c}
  \hline
  w/o TM & ECB & EMEA & Epps & GNOME & JRC & KDE & News & PHP & TED & Ubuntu & Wiki & All \\
  \hline
  copy & 59.8 & 64.5 & 34.4 & 70.3 & 67.6 & 55.3 & 12.0 & 38.6 & 30.8 & 51.6 & 47.4 & 52.6 \\
  \hline
  AR & 58.7 & 53.8 & 55.8 & 55.0 & 68.8 & 53.9 & 27.1 & 18.2 & 62.0 & 54.0 & 65.0 & 51.2 \\
  LevT & 46.6 & 30.7 & 51.8 & 51.0 & 62.3 & 47.0 & 23.6 & 12.5 & 58.7 & 50.0 & 61.9 & 46.5 \\
  \ourmodel & 53.0 & 49.7 & 53.2 & 51.5 & 64.7 & 50.8 & 24.5 & 37.1 & 59.5 & 50.4 & 64.0 & 52.6 \\
  \hline
  \hline
  w/ TM & ECB & EMEA & Epps & GNOME & JRC & KDE & News & PHP & TED & Ubuntu & Wiki & All \\
  \hline
  AR & 71.9 & 72.0 & 58.9 & 80.1 & 83.2 & 67.3 & 28.8 & 44.7 & 63.3 & 67.6 & 68.6 & 67.1 \\
  LevT & 62.4 & 53.8 & 55.5 & 77.5 & 78.8 & 63.3 & 26.1 & 28.7 & 60.2 & 66.0 & 67.1 & 60.4 \\
  ~~+tgt TM & 60.2 & 63.8 & 34.8 & 69.6 & 67.7 & 54.5 & 12.5 & 38.8 & 31.1 & 52.1 & 47.5 & 52.8 \\
  \ourmodel & 69.8 & 72.2 & 56.0 & 78.1 & 82.2 & 68.2 & 26.0 & 44.1 & 60.3 & 66.3 & 68.7 & 65.9 \\
  \hline
  \end{tabular}
  }
  \caption{BLEU scores for each domain when performing translation without and with TMs for $\operatorname{sim}>0.6$. \textit{All} is computed by concatenating all test sets ($11$k sentences in total). \textit{Copy} refers to using the TM match \simtrg{} as the output.\label{tab:fm-bleu}}
\end{table*}

We evaluate the performance of standard MT and TM-based MT on the two test sets ($\operatorname{sim}$$>$$0.6$ and $\operatorname{sim}$$\in$$[0.4, 0.6]$) introduced in Section~\ref{ssec:data}. When performing standard MT, the source side only contains the source sentence for all models, and the decoder side of \ourmodel{} is initialized with an empty input. When translating with TMs, the TM match is concatenated to the source sentence for all models. \ourmodel{} is additionally initialized with the TM match on the decoder side, a setting we also consider for LevT (+tgt TM). Table~\ref{tab:general-res} reports the aggregated results computed on all domains ($11$k sentences). The results for each domain when translating with and without TMs on $\operatorname{sim}$$>$$0.6$ are in Table~\ref{tab:fm-bleu}. The corresponding results with a breakdown by domain for $\operatorname{sim}$$\in$$[0.4, 0.6]$ are in Appendix~\ref{sec:res-detail}.

As reported in Tables~\ref{tab:general-res} and \ref{tab:fm-bleu}, the AR with TM baseline yields much higher BLEU and COMET scores than the standard MT setting. LevT can also make good use of TM matches, but its performance lags way behind the AR strategy in both settings.\footnote{We use the \texttt{fairseq} source code released by \citet{Gu19levenshtein} to train the LevT model. We have performed a sanity check by training a LevT model on the WMT14 English-German data and obtained results that are about $2$ BLEU points below the scores reported in \citet{Gu19levenshtein}. As \citet{Gu19levenshtein} have not specified the tool used to compute BLEU scores, it is difficult to make a more precise comparison.} Scores in Table~\ref{tab:fm-bleu} show that for domains like ECB and EMEA, it is difficult for LevT to generate good translations without using TMs, while for more general domains like News, the performance gap between LevT and AR are less significant. 

\begin{table}[!ht]
  \center
  \scalebox{0.85}{
  \begin{tabular}{l|cc|cc}
  \hline
        & \multicolumn{2}{c|}{$\operatorname{sim}>0.6$} & \multicolumn{2}{c}{$\operatorname{sim}\in[0.4, 0.6]$} \\
  \hline
  BLEU $\uparrow$ & w/o TM & w/ TM & w/o TM & w/ TM \\
  \hline
  copy & - & 52.6 & - & 34.5 \\
  \hline
  AR & 51.2 & 67.1 & 46.1 & 55.7 \\
  LevT & 46.5 & 60.4 & 40.8 & 49.3 \\
  $\quad$ +tgt TM & - & 52.8 & - & 35.0 \\
  \ourmodel & 52.6 & 65.9 & 45.7 & 53.3 \\
  \hline
  COMET $\uparrow$ & w/o TM & w/ TM & w/o TM & w/ TM \\
  \hline
  copy & - & 0.1330 & - & -0.3784 \\
  \hline
  AR & 0.6143 & 0.6985 & 0.5379 & 0.5900 \\
  LevT & 0.4251 & 0.5767 & 0.3429 & 0.4404 \\
  $\quad$ +tgt TM & - & 0.1639 & - & -0.3478 \\
  \ourmodel & 0.5314 & 0.6454 & 0.4263 & 0.4889 \\
  \hline
  \end{tabular}
  }
 \caption{BLEU and COMET scores on multi-domain test sets for various TM similarity ranges. \textit{w/o TM} is standard MT, \textit{w/ TM} adds a retrieved match \simtrg{} on the source side, and use it as initial target for \ourmodel. \textit{+tgt TM} refers to using TM match as the initial target for LevT.\label{tab:general-res}}
\end{table}

\ourmodel, on the contrary, does remarkably well when translating from scratch, even surpassing the AR model on BLEU on the $\operatorname{sim}$$>$$0.6$ set, which is arguably easier. When using TM matches, \ourmodel{} also performs much better than LevT. We achieve BLEU scores of only $1.2$ and $2.4$ below AR for $\operatorname{sim}$$>$$0.6$ and $\operatorname{sim}$$\in$$[0.4, 0.6]$, respectively (Table~\ref{tab:general-res}). The effect of using TM matches greatly varies across domains. \ourmodel{} can improve more general domains like News and Wiki when using TMs (Table~\ref{tab:fm-bleu}), even though the performance gains are less significant than for more specific domains like ECB and KDE. However, the same trend is also observed for AR, and \ourmodel{} even surpasses AR on Wiki in Table~\ref{tab:fm-bleu}. The gap in COMET score between \ourmodel{} and AR is also much smaller than reported by \citet{Helcl22nonautoregressive}, indicating that \ourmodel{} outputs valid translations. 

\paragraph{Unrelated Tokens} AR approaches are known to improperly copy "unrelated tokens" from TM matches into the output \citep{Xu20boosting}. As \ourmodel{} includes the deletion operation, we expect it to properly delete unrelated tokens. We define unrelated tokens as those present in \simtrg{} but not in \trg{} and count the ratio of such tokens that appear in the final translation. This is different from \citet{Xu20boosting}, as they used a word alignment model to label tokens that are unrelated to the "source". We have re-implemented the same method, but the alignment model was never perfect and yielded additional errors, which led to an imprecise measure of unrelated tokens. Table~\ref{tab:unrelated} reports the ratio of such unrelated tokens: \ourmodel{} is slightly less prone than AR to recopy unrelated parts of the TM matches in both test sets. LevT does even better on that account, but its comparatively lower BLEU scores suggest that it also discards valid tokens.

\begin{table}[!ht]
  \center
  \scalebox{0.85}{
  \begin{tabular}{l|cc}
  \hline
  Unrelated rate $\downarrow$ & $\operatorname{sim}>0.6$ & $\operatorname{sim}\in[0.4, 0.6]$ \\
  \hline
  AR & 28.42 & 17.78 \\
  LevT & 21.39 & 13.74 \\
  \ourmodel & 26.67 & 16.56 \\
  \hline
  \end{tabular}
  }
  \caption{Percentage of unrelated tokens from the retrieved TM matches appearing in the final translations.\label{tab:unrelated}}
\end{table}

\subsection{Ablation Analysis\label{ssec:ablation}}

\begin{table}[!ht]
  \center
  \scalebox{0.85}{
  \begin{tabular}{l|cc|cc}
  \hline
        & \multicolumn{2}{c|}{$\operatorname{sim}>0.6$} & \multicolumn{2}{c}{$\operatorname{sim}\in[0.4, 0.6]$} \\
  \hline
  BLEU & w/o TM & w/ TM & w/o TM & w/ TM \\
  \hline
  \ourmodel & 52.6 & 65.9 & 45.7 & 53.3 \\
  \hline
  $\quad$-tgt TM & 46.6 & 60.7 & 40.7 & 49.6 \\
  $\quad$-src TM & 53.2 & 64.3 & 45.9 & 52.2 \\
  \hline
  $\quad$-final-del & 38.5 & 64.2 & 32.7 & 50.8 \\
  $\quad$-self-pred & 52.6 & 65.2 & 45.6 & 52.7 \\
  \hline
  \end{tabular}
  }
  \caption{BLEU scores for various configurations. \textit{-tgt TM} (resp.\ \textit{-src TM}) is the model trained without TM match on the target (resp.\ source) side. \textit{-final-del} is trained without the final-del operation, \textit{-self-pred} only applies reference deletions during training.\label{tab:ablation}}
\end{table}

We conduct an ablation analysis to study the effectiveness of each component of our method, by training a new model for each contrast. Training without TM matches on the target side (Table~\ref{tab:ablation}, -tgt TM) vastly degrades the performance in both conditions (w/ and w/o TM), indicating that the standard MT can also benefit from training with TMs as initial targets. However, removing TM matches on the source side (-src TM) improves standard MT, as also observed by \citet{Bulte19fuzzy}, but has a negative impact when translating with TMs. This highlights the importance of always remembering the TM match on the source side of \ourmodel. 

We also compare with alternative implementations of the deletion operation. Results in Table~\ref{tab:ablation} (-final-del) show that removing the final deletion step mostly impacts \ourmodel{} in the standard MT setting, where the detection of wrong predictions matters most \cite{Huang22improving}. This further demonstrates that unrelated tokens from TM matches and the prediction errors of the model are vastly different, and training to delete both is necessary. Last, we experiment with using only reference deletion labels to train the insertion operation, instead of using both the reference and model predictions (see Section~\ref{ssec:tmlevt}). We observe (-self-pred) a small performance drop with respect to the baseline policy.

\subsection{Knowledge Distillation\label{ssec:distillation}}

KD is used in most NAT models, as it reduces the complexity and lexical diversity of target sentences, thereby helping NAT approaches to mitigate the multimodality issue \citep{Zhou20understanding,Xu21distilled}. Given that our results so far have only relied on actual target data, we thus ask whether KD could also improve the performance of TM-based NAT. We train a teacher Transformer-based model with the $4.4$M parallel data and use it for data distillation. As expected, using KD does improve BLEU scores of \ourmodel{} on standard MT (Table~\ref{tab:distillation}). However, using KD hurts performance when editing an initial similar translation, resulting in a large drop in scores compared to using real data. Applying KD also to TM matches yields similar results.

\begin{table}[!ht]
  \center
  \scalebox{0.85}{
  \begin{tabular}{l|cc|cc}
  \hline
        & \multicolumn{2}{c|}{$\operatorname{sim}>0.6$} & \multicolumn{2}{c}{$\operatorname{sim}\in[0.4, 0.6]$} \\
  \hline
  BLEU & w/o TM & w/ TM & w/o TM & w/ TM \\
  \hline
  copy & - & 52.6 & - & 34.5 \\
  \hline
  Teacher & 56.7 & - & 49.6 & - \\
  \hline
  AR + KD & 55.7 & 56.9 & 48.7 & 49.4 \\
  \ourmodel & 52.6 & 65.9 & 45.7 & 53.3 \\
  $\quad$+KD & 54.3 & 57.1 & 47.6 & 49.3 \\
  $\quad$+KD TM & 53.8 & 56.0 & 47.3 & 48.5 \\
  \hline
  \end{tabular}
  }
  \caption{BLEU scores with and without KD. \textit{Teacher} is the teacher model used to distill the parallel data. \textit{+KD} applies KD to the training references, \textit{+KD TM} applies KD to both references and TM matches.\label{tab:distillation}}
\end{table}

The benefits of KD are assumed to mainly reduce the multimodality issue in NAT \citep{Zhou20understanding}. This issue may be less problematic in our context, as the TM match already provides an explicit and often unambiguous context for generating the missing words in the translation. In this case, KD is even detrimental to the translation quality. This is because using distilled references exposes \ourmodel{} to imperfect translations (with BLEU scores of respectively $56.7$ and $49.6$), only a few points better than the initial TM matches (\textit{copy} in Table~\ref{tab:distillation}). This seems to limit \ourmodel's ability in learning to generate very high quality translation that it can achieve when exposed to real references (+$8.8$ and +$4$ BLEU on $\operatorname{sim}$$>$$0.6$ and $\operatorname{sim}$$\in$$[0.4, 0.6]$, respectively). For comparison, we also train an AR model using the same KD data (AR + KD) and again observe very little gain when translating with TMs. In fact, the limit of models using KD data, whether using TMs or not, seems to be upper-bounded by the performance of the teacher. These intuitions are illustrated by the example in Figure~\ref{fig:example}.

\begin{figure*}[ht]
  \center
  \scalebox{0.83}{
  \begin{tabular}{l|l}
  \hline
  SRC & Measures to reduce or eliminate releases from unintentional production \\
  \hline
  TM match & Measures to reduce or eliminate releases from intentional production and use \\
  $\operatorname{sim}$$=$$0.73$ & Mesures propres à réduire ou éliminer les rejets résultant d'une production et d'une utilisation intentionnelles \\
  \hline
  \ourmodel & Mesures visant à réduire ou à éliminer les disséminations de la production non intentionnelle \\
  $\quad$+TM & Mesures \textcolor{blue}{\textbf{propres}} à réduire \textcolor{blue}{\textbf{ou éliminer}} les \textcolor{blue}{\textbf{rejets résultant d'une}} production non intentionnelle \\
  \hline
  KD & Mesures visant à réduire ou à éliminer les rejets de la production non intentionnelle \\
  $\quad$+TM & Mesures \textcolor{red}{\textbf{visant}} à réduire ou \textcolor{red}{\textbf{à}} éliminer les rejets \textcolor{red}{\textbf{provenant de la}} production non intentionnelle \\
  \hline
  REF & Mesures propres à réduire ou éliminer les rejets résultant d'une production non intentionnelle \\
  \hline
  \end{tabular}
  }
  \caption{An example with the retrieved TM match and translations generated by \ourmodel{} and KD model. Benefits taken from the TM match by \ourmodel{} are in blue. Segments KD model fails to make use of are in red.\label{fig:example}}
\end{figure*}

\subsection{Computational Trade-offs of \ourmodel\label{ssec:efficiency}}

Inference speedup is the main advantage of using NAT models. Table~\ref{tab:time} compares the average decoding time per sentence on all domains.\footnote{We exclude PHP, for which AR generates many repetitions, yielding very long runtimes that biased the average.} Here, we perform the inference speed analysis as a sanity check. As discussed by \citet{Kasai2021deep}, \citet{Helcl22nonautoregressive} and \citet{Schmidt22nonautoregressive}, comparing the inference speed for NAT models could be tricky. We follow here the recommendations of \citet{Helcl22nonautoregressive} and use the same hardware conditions and inference batch size for all settings, making our results as comparable as possible. \ourmodel{} is much faster than AR both with and without TMs. We also note that decoding with a TM match is always slightly longer due to (a) finding matches; (b) encoding a longer input  made of the source and the TM match.\footnote{The numbers in Table~\ref{tab:time} only reflect the effect of (b), as step (a) is the same for both AR and NAT models.}

\begin{table}[!ht]
  \center
  \scalebox{0.85}{
  \begin{tabular}{l|cc|c}
  \hline
  Settings & AR & \ourmodel & Speedup \\
  \hline
  w/o TM & 5.91 & 2.53 & $\times$2.34 \\
  w/ TM & 7.80 & 3.43 & $\times$2.28 \\
  \hline
  \end{tabular}
  }
  \caption{Average decoding time (ms) per sentence for all domains of both $\operatorname{sim}>0.6$ and $\operatorname{sim}\in[0.4, 0.6]$.\label{tab:time}}
\end{table}

\begin{table}[!ht]
  \center
  \scalebox{0.85}{
  \begin{tabular}{l|cc|cc}
  \hline
        & \multicolumn{2}{c|}{$\operatorname{sim}>0.6$} & \multicolumn{2}{c}{$\operatorname{sim}\in[0.4, 0.6]$} \\
  \hline 
   Systems & w/o TM & w/ TM & w/o TM & w/ TM \\
  \hline
  LevT & 2.027 & 1.899 & 2.544 & 2.538 \\
  \ourmodel & 1.781 & 1.348 & 2.260 & 1.880 \\
  \hline
  \end{tabular}
  }
  \caption{Average number of decoding iterations.\label{tab:efficiency}}
\end{table}

Using TMs in MT is expected to improve the translation quality while also reducing the decoding load, as useful tokens can be directly copied to the output. This is not observed in AR approaches nor in the original LevT, as both models always start inference with an empty input. \ourmodel, however, uses an initial translation to speed up decoding. We report the average number of iterations required in inference in Table~\ref{tab:efficiency}, where we see that translating with TMs reduces the decoding effort for \ourmodel{} by about $20\%$, while it remains almost unchanged for LevT. We also find that \ourmodel{} needs fewer iterations to converge than LevT in all conditions. Note that the training time of \ourmodel{} is only $1.1$-$1.2$$\times$ compared to LevT, which we think is an acceptable overhead considering the large performance gains and the reduction of decoding load. 

\subsection{Good Matches Increase MT Quality\label{ssec:threshold}}

\begin{figure}[ht]
  \center
  \includegraphics[width=\columnwidth]{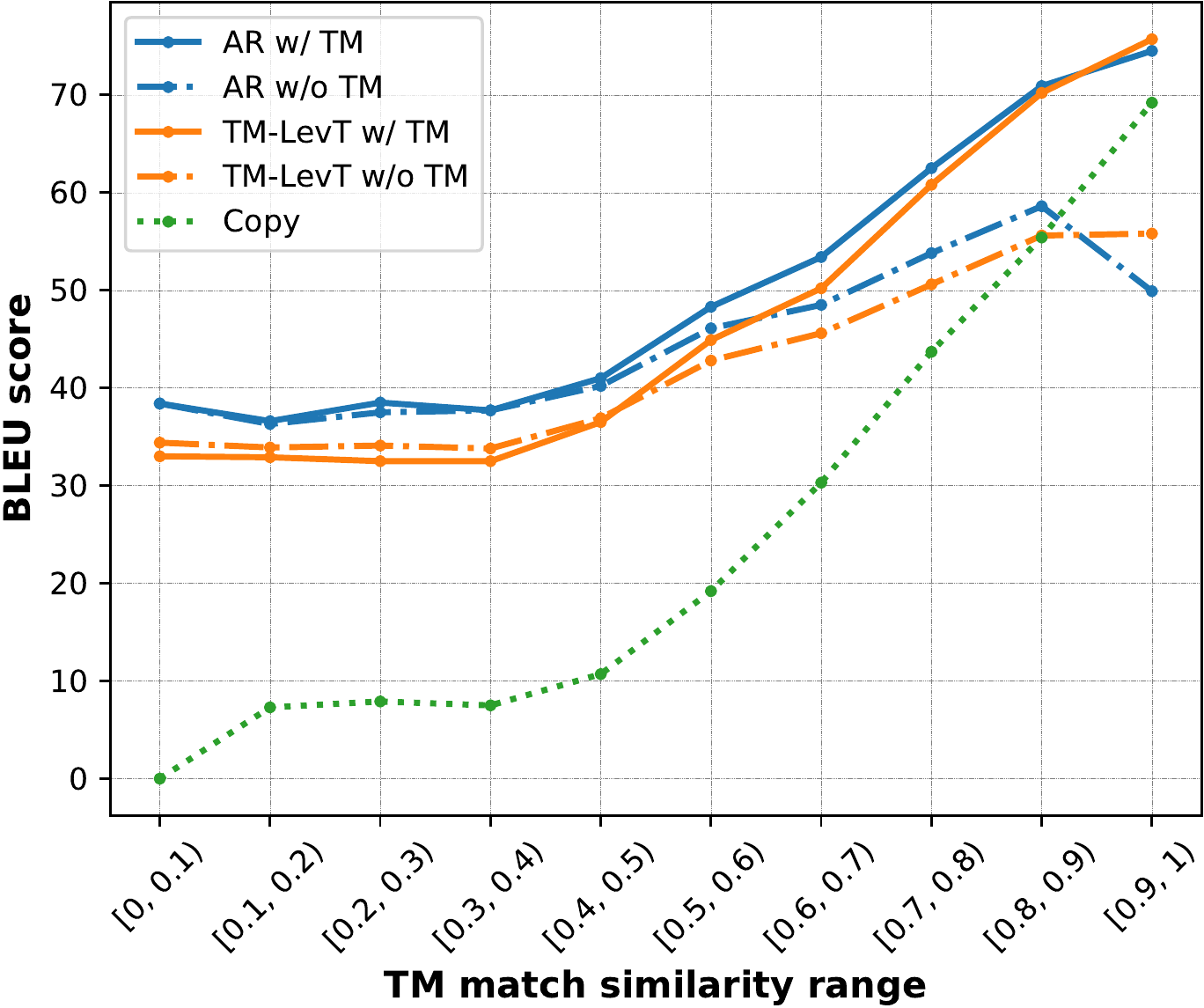}
  \caption{BLEU scores when translating with TM matches of various similarity ranges. The \textit{Copy} strategy uses the retrieved example as the final translation.\label{fig:threshold}}
\end{figure}

TM-based models require good TM matches to improve their translations; when none is found, standard MT can be used instead. Defining the minimal similarity for a match to be useful ($0.4$ in this paper) requires some tuning and the best threshold may vary from corpora to corpora. In this section, we take a closer look at the effect of thresholding for the AR and NAT models considered here. We compute the best TM match for the held-out set of Section~\ref{ssec:data}, trying to find a matched translation for all sentences without any filtering. We then combine the held-out set with the two test sets ($\operatorname{sim}$$>$$0.6$ and $\operatorname{sim}$$\in$$[0.4, 0.6]$) and bucket sentences by the similarity of the best TM match. For each bucket, we compute BLEU scores obtained for various systems and plot results on Figure~\ref{fig:threshold}.

We first observe that the scores of all systems, even those that do not use TMs, increase with better TM matches: this is because test sentences with better matches are also more similar to the training data, thus easier to translate for all systems. Secondly, even with very good matches, both models using TMs are able to improve over the \textit{Copy} strategy. Comparing AR to \ourmodel, we see that the former is preferable across the board, even though the gap closes when very good matches are available. The TM-based AR model is almost as good as standard MT for poor matches and starts improving standard MT when $\operatorname{sim}$$\geq$$0.4$. In comparison, we see for \ourmodel{} a small edge of standard MT over translating with TMs which subsists as long as $\operatorname{sim}$$<$$0.5$; for higher similarity matches, translating with TMs gets much better scores. These results suggest that thresholding may not be necessary and that both TM-based architectures adapt their behavior to the match quality, with hardly any performance loss w.r.t.\ the standard MT approach.

\section{Related Work\label{sec:related}}

\paragraph{TM-based MT} Most studies using TM sentences to improve translations are based on AR models and either use a second encoder to integrate the TM match or concatenate it to the source in the same encoder. The former approach is illustrated by \citet{Gu18search}, who rerun the translation model as an encoder to encode the similar translation. \citet{Xia19graphbased} alternatively use a compact graph as the second encoder. Another work along this line is \citet{He21fast}, which encodes the TM match using the decoder embedding matrix and performs a cross-attention between the decoder input and the encoded TM match. Besides, \citet{Cai21neural} directly search in a corpus of target sentences with a cross-lingual similarity and encode the resulting sentences with a dual encoder approach similar to \citet{Junczys-dowmunt18ms}. Single encoder approaches are first explored by \citet{Bulte19fuzzy}, who concatenate the TM match with the source to perform TM-based MT. This idea is extended by \citet{Xu20boosting} by adding a second embedding feature to distinguish related and unrelated tokens and by \citet{Pham20priming}, who use both source and target sentences of the matched TM. \citet{Zhang18guiding} explore a different direction to improve translation with retrieved segments instead of complete sentences. \citet{Khandelwal21nearest} further propose $k$-nearest neighbor MT by searching for target tokens that have similar contextualized representations at each decoding step, an approach continued by \citet{Zheng21adaptive} with dynamic neighborhoods.

\paragraph{NAT with Augmented Resources} Several works have studied ways to integrate extra information into NAT architectures, mostly using the LevT model as their starting point. \citet{Susanto20lexically} incorporate lexical constraints with LevT by simply initializing the decoder with constraint words inserted in a predefined order; this limitation is lifted in the EDITOR model of \citet{Xu21editor}, who introduce a repositioning operation to allow constraints to be inserted in arbitrary order. \citet{Zeng22neighbors} pay particular attention to low-frequency constraints by preventing rare tokens from being removed when generating training samples for the insertion operation. The most relevant study to this paper is the recent work by \citet{Niwa22nearest}, who also seek to improve LevT with TMs, using good matches to initialize the decoder. This work only mildly departs from the vanilla LevT with a small modification of the deletion operation to remove unrelated tokens and only compare with standard MT, failing to contrast their improvements with TM-based AR models. Finally, \citet{Xu22bilingual} also explore the integration of TMs into the original LevT model, but fail to obtain improvements over a copy baseline.

\section{Conclusion\label{sec:conclusion}}

In this paper, we studied ways to augment the LevT architecture with TMs. Our proposal adds an initial deletion operation during training to detect possible unrelated tokens present in TM matches. By copying the TM match both on the source side and on the target side as an initial target sequence, our model vastly outperformed the original LevT model and achieved BLEU scores approaching those of a strong AR model both when decoding from scratch and when editing a TM match. Compared to LevT, \ourmodel{} also generates translations that contain less unrelated tokens, and is able to converge in fewer iterations. We also found that training with TMs improves NAT performance on standard MT. Finally, we have tried to combine KD with our approach, concluding that it was more hurting than helpful for TM-based architectures.

\section*{Limitations}

NAT models such as LevT are more difficult to train than AR models, as they require larger batch size to converge. Our \ourmodel{} adds an initial deletion operation during training, therefore slightly lengthening the training time by approximately $1.1-1.2\times$ with respect to the basic LevT model. Due to computational limits, we have not conducted experiments on other language pairs, especially on more distant language pairs. Even tough our findings apply for a wide range of domains, considering also more languages would be helpful to fully validate our observations.

\section*{Acknowledgement}

The authors would like to thank Maxime Bouthors for his suggestions and Ruiyang Zhou for her help of the deficiency analysis of LevT. We would also like to thank the anonymous reviewers for their valuable suggestions and comments. This work was granted access to the HPC resources of IDRIS under the allocation 2022-[AD011011580R2] made by GENCI. The first author is partly funded by SYSTRAN and by a grant (Transwrite) from the Région Ile-de-France.

\bibliography{biblio}

\begin{thebibliography}{39}
\expandafter\ifx\csname natexlab\endcsname\relax\def\natexlab#1{#1}\fi

\bibitem[{Bulte and Tezcan(2019)}]{Bulte19fuzzy}
Bram Bulte and Arda Tezcan. 2019.
\newblock \href {https://doi.org/10.18653/v1/P19-1175} {Neural fuzzy repair:
  Integrating fuzzy matches into neural machine translation}.
\newblock In \emph{Proceedings of the 57th Annual Meeting of the Association
  for Computational Linguistics}, pages 1800--1809, Florence, Italy.
  Association for Computational Linguistics.

\bibitem[{Cai et~al.(2021)Cai, Wang, Li, Lam, and Liu}]{Cai21neural}
Deng Cai, Yan Wang, Huayang Li, Wai Lam, and Lemao Liu. 2021.
\newblock \href {https://doi.org/10.18653/v1/2021.acl-long.567} {Neural machine
  translation with monolingual translation memory}.
\newblock In \emph{Proceedings of the 59th Annual Meeting of the Association
  for Computational Linguistics and the 11th International Joint Conference on
  Natural Language Processing (Volume 1: Long Papers)}, pages 7307--7318,
  Online. Association for Computational Linguistics.

\bibitem[{Ghazvininejad et~al.(2019)Ghazvininejad, Levy, Liu, and
  Zettlemoyer}]{Ghazvininejad19mask}
Marjan Ghazvininejad, Omer Levy, Yinhan Liu, and Luke Zettlemoyer. 2019.
\newblock \href {https://doi.org/10.18653/v1/D19-1633} {Mask-predict: Parallel
  decoding of conditional masked language models}.
\newblock In \emph{Proceedings of the 2019 Conference on Empirical Methods in
  Natural Language Processing and the 9th International Joint Conference on
  Natural Language Processing (EMNLP-IJCNLP)}, pages 6112--6121, Hong Kong,
  China. Association for Computational Linguistics.

\bibitem[{Gu et~al.(2018{\natexlab{a}})Gu, Bradbury, Xiong, Li, and
  Socher}]{Gu18nonautoregressive}
Jiatao Gu, James Bradbury, Caiming Xiong, Victor~O.K. Li, and Richard Socher.
  2018{\natexlab{a}}.
\newblock \href {https://openreview.net/forum?id=B1l8BtlCb} {Non-autoregressive
  neural machine translation}.
\newblock In \emph{International Conference on Learning Representations}.

\bibitem[{Gu and Kong(2021)}]{Gu21fully}
Jiatao Gu and Xiang Kong. 2021.
\newblock \href {https://doi.org/10.18653/v1/2021.findings-acl.11} {Fully
  non-autoregressive neural machine translation: Tricks of the trade}.
\newblock In \emph{Findings of the Association for Computational Linguistics:
  ACL-IJCNLP 2021}, pages 120--133, Online. Association for Computational
  Linguistics.

\bibitem[{Gu et~al.(2019)Gu, Wang, and Zhao}]{Gu19levenshtein}
Jiatao Gu, Changhan Wang, and Junbo Zhao. 2019.
\newblock \href
  {https://proceedings.neurips.cc/paper/2019/file/675f9820626f5bc0afb47b57890b466e-Paper.pdf}
  {Levenshtein transformer}.
\newblock In \emph{Advances in Neural Information Processing Systems},
  volume~32, pages 11181--11191. Curran Associates, Inc.

\bibitem[{Gu et~al.(2018{\natexlab{b}})Gu, Wang, Cho, and Li}]{Gu18search}
Jiatao Gu, Yong Wang, Kyunghyun Cho, and Victor~O.K. Li. 2018{\natexlab{b}}.
\newblock \href {https://doi.org/10.1609/aaai.v32i1.12013} {Search engine
  guided neural machine translation}.
\newblock \emph{Proceedings of the AAAI Conference on Artificial Intelligence},
  32(1).

\bibitem[{He et~al.(2021)He, Huang, Cui, Li, and Liu}]{He21fast}
Qiuxiang He, Guoping Huang, Qu~Cui, Li~Li, and Lemao Liu. 2021.
\newblock \href {https://doi.org/10.18653/v1/2021.acl-long.246} {Fast and
  accurate neural machine translation with translation memory}.
\newblock In \emph{Proceedings of the 59th Annual Meeting of the Association
  for Computational Linguistics and the 11th International Joint Conference on
  Natural Language Processing (Volume 1: Long Papers)}, pages 3170--3180,
  Online. Association for Computational Linguistics.

\bibitem[{Helcl et~al.(2022)Helcl, Haddow, and
  Birch}]{Helcl22nonautoregressive}
Jind{\v{r}}ich Helcl, Barry Haddow, and Alexandra Birch. 2022.
\newblock \href {https://aclanthology.org/2022.naacl-main.129}
  {Non-autoregressive machine translation: It{'}s not as fast as it seems}.
\newblock In \emph{Proceedings of the 2022 Conference of the North American
  Chapter of the Association for Computational Linguistics: Human Language
  Technologies}, pages 1780--1790, Seattle, United States. Association for
  Computational Linguistics.

\bibitem[{Huang et~al.(2022)Huang, Perez, and Volkovs}]{Huang22improving}
Xiao~Shi Huang, Felipe Perez, and Maksims Volkovs. 2022.
\newblock \href {https://openreview.net/forum?id=I2Hw58KHp8O} {Improving
  non-autoregressive translation models without distillation}.
\newblock In \emph{International Conference on Learning Representations}.

\bibitem[{Junczys-Dowmunt and Grundkiewicz(2018)}]{Junczys-dowmunt18ms}
Marcin Junczys-Dowmunt and Roman Grundkiewicz. 2018.
\newblock \href {https://doi.org/10.18653/v1/W18-6467} {{MS}-{UE}din submission
  to the {WMT}2018 {APE} shared task: Dual-source transformer for automatic
  post-editing}.
\newblock In \emph{Proceedings of the Third Conference on Machine Translation:
  Shared Task Papers}, pages 822--826, Belgium, Brussels. Association for
  Computational Linguistics.

\bibitem[{Kasai et~al.(2021)Kasai, Pappas, Peng, Cross, and
  Smith}]{Kasai2021deep}
Jungo Kasai, Nikolaos Pappas, Hao Peng, James Cross, and Noah Smith. 2021.
\newblock \href {https://openreview.net/forum?id=KpfasTaLUpq} {Deep encoder,
  shallow decoder: Reevaluating non-autoregressive machine translation}.
\newblock In \emph{International Conference on Learning Representations}.

\bibitem[{Khandelwal et~al.(2021)Khandelwal, Fan, Jurafsky, Zettlemoyer, and
  Lewis}]{Khandelwal21nearest}
Urvashi Khandelwal, Angela Fan, Dan Jurafsky, Luke Zettlemoyer, and Mike Lewis.
  2021.
\newblock \href {https://openreview.net/forum?id=7wCBOfJ8hJM} {Nearest neighbor
  machine translation}.
\newblock In \emph{International Conference on Learning Representations}.

\bibitem[{Kim and Rush(2016)}]{Kim16sequence}
Yoon Kim and Alexander~M. Rush. 2016.
\newblock \href {https://doi.org/10.18653/v1/D16-1139} {Sequence-level
  knowledge distillation}.
\newblock In \emph{Proceedings of the 2016 Conference on Empirical Methods in
  Natural Language Processing}, pages 1317--1327, Austin, Texas. Association
  for Computational Linguistics.

\bibitem[{Koehn and Senellart(2010)}]{Koehn10convergence}
Philipp Koehn and Jean Senellart. 2010.
\newblock \href {https://aclanthology.org/2010.jec-1.4} {Convergence of
  translation memory and statistical machine translation}.
\newblock In \emph{Proceedings of the Second Joint EM+/CNGL Workshop: Bringing
  MT to the User: Research on Integrating MT in the Translation Industry},
  pages 21--32, Denver, Colorado, USA. Association for Machine Translation in
  the Americas.

\bibitem[{Libovick{\'y} and Helcl(2018)}]{Libovicky18end}
Jind{\v{r}}ich Libovick{\'y} and Jind{\v{r}}ich Helcl. 2018.
\newblock \href {https://doi.org/10.18653/v1/D18-1336} {End-to-end
  non-autoregressive neural machine translation with connectionist temporal
  classification}.
\newblock In \emph{Proceedings of the 2018 Conference on Empirical Methods in
  Natural Language Processing}, pages 3016--3021, Brussels, Belgium.
  Association for Computational Linguistics.

\bibitem[{Niwa et~al.(2022)Niwa, Takase, and Okazaki}]{Niwa22nearest}
Ayana Niwa, Sho Takase, and Naoaki Okazaki. 2022.
\newblock \href {https://doi.org/10.48550/ARXIV.2208.12496} {Nearest neighbor
  non-autoregressive text generation}.
\newblock \emph{CoRR}, abs/2208.12496.

\bibitem[{Ott et~al.(2019)Ott, Edunov, Baevski, Fan, Gross, Ng, Grangier, and
  Auli}]{Ott19fairseq}
Myle Ott, Sergey Edunov, Alexei Baevski, Angela Fan, Sam Gross, Nathan Ng,
  David Grangier, and Michael Auli. 2019.
\newblock \href {https://doi.org/10.18653/v1/N19-4009} {fairseq: A fast,
  extensible toolkit for sequence modeling}.
\newblock In \emph{Proceedings of the 2019 Conference of the North {A}merican
  Chapter of the Association for Computational Linguistics (Demonstrations)},
  pages 48--53, Minneapolis, Minnesota. Association for Computational
  Linguistics.

\bibitem[{Pham et~al.(2020)Pham, Xu, Crego, Yvon, and
  Senellart}]{Pham20priming}
Minh~Quang Pham, Jitao Xu, Josep Crego, Fran{\c c}ois Yvon, and Jean Senellart.
  2020.
\newblock \href {https://www.aclweb.org/anthology/2020.wmt-1.63} {Priming
  neural machine translation}.
\newblock In \emph{Proceedings of the Fifth Conference on Machine Translation},
  pages 462--473, Online. Association for Computational Linguistics.

\bibitem[{Post(2018)}]{Post18sacrebleu}
Matt Post. 2018.
\newblock \href {https://doi.org/10.18653/v1/W18-6319} {A call for clarity in
  reporting {BLEU} scores}.
\newblock In \emph{Proceedings of the Third Conference on Machine Translation:
  Research Papers}, pages 186--191, Brussels, Belgium. Association for
  Computational Linguistics.

\bibitem[{Press and Wolf(2017)}]{Press17using}
Ofir Press and Lior Wolf. 2017.
\newblock \href {https://aclanthology.org/E17-2025} {Using the output embedding
  to improve language models}.
\newblock In \emph{Proceedings of the 15th Conference of the {E}uropean Chapter
  of the Association for Computational Linguistics: Volume 2, Short Papers},
  pages 157--163, Valencia, Spain. Association for Computational Linguistics.

\bibitem[{Rei et~al.(2020)Rei, Stewart, Farinha, and Lavie}]{Rei20comet}
Ricardo Rei, Craig Stewart, Ana~C Farinha, and Alon Lavie. 2020.
\newblock \href {https://doi.org/10.18653/v1/2020.emnlp-main.213} {{COMET}: A
  neural framework for {MT} evaluation}.
\newblock In \emph{Proceedings of the 2020 Conference on Empirical Methods in
  Natural Language Processing (EMNLP)}, pages 2685--2702, Online. Association
  for Computational Linguistics.

\bibitem[{Saharia et~al.(2020)Saharia, Chan, Saxena, and
  Norouzi}]{Saharia20nonauto}
Chitwan Saharia, William Chan, Saurabh Saxena, and Mohammad Norouzi. 2020.
\newblock \href {https://www.aclweb.org/anthology/2020.emnlp-main.83}
  {Non-autoregressive machine translation with latent alignments}.
\newblock In \emph{Proceedings of the 2020 Conference on Empirical Methods in
  Natural Language Processing (EMNLP)}, pages 1098--1108, Online. Association
  for Computational Linguistics.

\bibitem[{Schmidt et~al.(2022)Schmidt, Pires, Peitz, and
  L{\"o}{\"o}f}]{Schmidt22nonautoregressive}
Robin Schmidt, Telmo Pires, Stephan Peitz, and Jonas L{\"o}{\"o}f. 2022.
\newblock \href {https://aclanthology.org/2022.emnlp-main.179}
  {Non-autoregressive neural machine translation: A call for clarity}.
\newblock In \emph{Proceedings of the 2022 Conference on Empirical Methods in
  Natural Language Processing}, pages 2785--2799, Abu Dhabi, United Arab
  Emirates. Association for Computational Linguistics.

\bibitem[{Sennrich et~al.(2016)Sennrich, Haddow, and Birch}]{Sennrich16BPE}
Rico Sennrich, Barry Haddow, and Alexandra Birch. 2016.
\newblock \href {https://doi.org/10.18653/v1/P16-1162} {Neural machine
  translation of rare words with subword units}.
\newblock In \emph{Proceedings of the 54th Annual Meeting of the Association
  for Computational Linguistics (Volume 1: Long Papers)}, pages 1715--1725,
  Berlin, Germany. Association for Computational Linguistics.

\bibitem[{Susanto et~al.(2020)Susanto, Chollampatt, and
  Tan}]{Susanto20lexically}
Raymond~Hendy Susanto, Shamil Chollampatt, and Liling Tan. 2020.
\newblock \href {https://doi.org/10.18653/v1/2020.acl-main.325} {Lexically
  constrained neural machine translation with {L}evenshtein transformer}.
\newblock In \emph{Proceedings of the 58th Annual Meeting of the Association
  for Computational Linguistics}, pages 3536--3543, Online. Association for
  Computational Linguistics.

\bibitem[{Tiedemann(2012)}]{Tiedemann12parallel}
J{\"o}rg Tiedemann. 2012.
\newblock \href
  {http://www.lrec-conf.org/proceedings/lrec2012/pdf/463_Paper.pdf} {Parallel
  data, tools and interfaces in {OPUS}}.
\newblock In \emph{Proceedings of the Eighth International Conference on
  Language Resources and Evaluation ({LREC}'12)}, pages 2214--2218, Istanbul,
  Turkey. European Language Resources Association (ELRA).

\bibitem[{Vaswani et~al.(2017)Vaswani, Shazeer, Parmar, Uszkoreit, Jones,
  Gomez, Kaiser, and Polosukhin}]{Vaswani17attention}
Ashish Vaswani, Noam Shazeer, Niki Parmar, Jakob Uszkoreit, Llion Jones,
  Aidan~N Gomez, \L~ukasz Kaiser, and Illia Polosukhin. 2017.
\newblock \href
  {http://papers.nips.cc/paper/7181-attention-is-all-you-need.pdf} {Attention
  is all you need}.
\newblock In I.~Guyon, U.~V. Luxburg, S.~Bengio, H.~Wallach, R.~Fergus,
  S.~Vishwanathan, and R.~Garnett, editors, \emph{Advances in Neural
  Information Processing Systems 30}, pages 5998--6008. Curran Associates, Inc.

\bibitem[{Xia et~al.(2019)Xia, Huang, Liu, and Shi}]{Xia19graphbased}
Mengzhou Xia, Guoping Huang, Lemao Liu, and Shuming Shi. 2019.
\newblock \href {https://doi.org/10.1609/aaai.v33i01.33017297} {Graph based
  translation memory for neural machine translation}.
\newblock \emph{Proceedings of the AAAI Conference on Artificial Intelligence},
  33(01):7297--7304.

\bibitem[{Xiao et~al.(2022)Xiao, Wu, Guo, Li, Zhang, Qin, and
  Liu}]{Xiao22surveyNAT}
Yisheng Xiao, Lijun Wu, Junliang Guo, Juntao Li, Min Zhang, Tao Qin, and
  Tie-yan Liu. 2022.
\newblock \href {https://doi.org/10.48550/ARXIV.2204.09269} {A survey on
  non-autoregressive generation for neural machine translation and beyond}.
\newblock \emph{CoRR}, abs/2204.09269.

\bibitem[{Xu et~al.(2020)Xu, Crego, and Senellart}]{Xu20boosting}
Jitao Xu, Josep Crego, and Jean Senellart. 2020.
\newblock \href {https://doi.org/10.18653/v1/2020.acl-main.144} {Boosting
  neural machine translation with similar translations}.
\newblock In \emph{Proceedings of the 58th Annual Meeting of the Association
  for Computational Linguistics}, pages 1580--1590, Online. Association for
  Computational Linguistics.

\bibitem[{Xu et~al.(2022)Xu, Crego, and Yvon}]{Xu22bilingual}
Jitao Xu, Josep Crego, and Fran{\c{c}}ois Yvon. 2022.
\newblock \href {https://aclanthology.org/2022.emnlp-main.548} {Bilingual
  synchronization: Restoring translational relationships with editing
  operations}.
\newblock In \emph{Proceedings of the 2022 Conference on Empirical Methods in
  Natural Language Processing}, pages 8016--8030, Abu Dhabi, United Arab
  Emirates. Association for Computational Linguistics.

\bibitem[{Xu and Carpuat(2021)}]{Xu21editor}
Weijia Xu and Marine Carpuat. 2021.
\newblock \href {https://doi.org/10.1162/tacl_a_00368} {{EDITOR: An Edit-Based
  Transformer with Repositioning for Neural Machine Translation with Soft
  Lexical Constraints}}.
\newblock \emph{Transactions of the Association for Computational Linguistics},
  9:311--328.

\bibitem[{Xu et~al.(2021)Xu, Ma, Zhang, and Carpuat}]{Xu21distilled}
Weijia Xu, Shuming Ma, Dongdong Zhang, and Marine Carpuat. 2021.
\newblock \href {https://doi.org/10.18653/v1/2021.findings-acl.385} {How does
  distilled data complexity impact the quality and confidence of
  non-autoregressive machine translation?}
\newblock In \emph{Findings of the Association for Computational Linguistics:
  ACL-IJCNLP 2021}, pages 4392--4400, Online. Association for Computational
  Linguistics.

\bibitem[{Yamada(2011)}]{Yamada11effect}
Masaru Yamada. 2011.
\newblock \href
  {http://www.intercultural.urv.cat/media/upload/domain_317/arxius/TP3/isgbook3_web.pdf\#page=55}
  {The effect of translation memory databases on productivity}.
\newblock \emph{Translation research projects}, 3:63--73.

\bibitem[{Zeng et~al.(2022)Zeng, Chen, Zhuang, Xu, Yang, Ying, Tao, and
  Xiao}]{Zeng22neighbors}
Chun Zeng, Jiangjie Chen, Tianyi Zhuang, Rui Xu, Hao Yang, Qin Ying, Shimin
  Tao, and Yanghua Xiao. 2022.
\newblock \href {https://doi.org/10.18653/v1/2022.naacl-main.424} {Neighbors
  are not strangers: Improving non-autoregressive translation under
  low-frequency lexical constraints}.
\newblock In \emph{Proceedings of the 2022 Conference of the North American
  Chapter of the Association for Computational Linguistics: Human Language
  Technologies}, pages 5777--5790, Seattle, United States. Association for
  Computational Linguistics.

\bibitem[{Zhang et~al.(2018)Zhang, Utiyama, Sumita, Neubig, and
  Nakamura}]{Zhang18guiding}
Jingyi Zhang, Masao Utiyama, Eiichro Sumita, Graham Neubig, and Satoshi
  Nakamura. 2018.
\newblock \href {https://doi.org/10.18653/v1/N18-1120} {Guiding neural machine
  translation with retrieved translation pieces}.
\newblock In \emph{Proceedings of the 2018 Conference of the North {A}merican
  Chapter of the Association for Computational Linguistics: Human Language
  Technologies, Volume 1 (Long Papers)}, pages 1325--1335, New Orleans,
  Louisiana. Association for Computational Linguistics.

\bibitem[{Zheng et~al.(2021)Zheng, Zhang, Guo, Huang, Chen, Luo, and
  Chen}]{Zheng21adaptive}
Xin Zheng, Zhirui Zhang, Junliang Guo, Shujian Huang, Boxing Chen, Weihua Luo,
  and Jiajun Chen. 2021.
\newblock \href {https://doi.org/10.18653/v1/2021.acl-short.47} {Adaptive
  nearest neighbor machine translation}.
\newblock In \emph{Proceedings of the 59th Annual Meeting of the Association
  for Computational Linguistics and the 11th International Joint Conference on
  Natural Language Processing (Volume 2: Short Papers)}, pages 368--374,
  Online. Association for Computational Linguistics.

\bibitem[{Zhou et~al.(2020)Zhou, Gu, and Neubig}]{Zhou20understanding}
Chunting Zhou, Jiatao Gu, and Graham Neubig. 2020.
\newblock \href {https://openreview.net/forum?id=BygFVAEKDH} {Understanding
  knowledge distillation in non-autoregressive machine translation}.
\newblock In \emph{International Conference on Learning Representations}.

\end{thebibliography}
\bibliographystyle{acl_natbib}

\clearpage
\appendix

\section{Details of Data Processing\label{sec:data}}

Table~\ref{tab:data-stat} reports statistics of the ratio of TM matches for various similarity ranges of the multi-domain dataset described in Section~\ref{ssec:data}. These ratios vary greatly across domains. 

\begin{table}[!ht]
  \center
  \scalebox{0.85}{
  \begin{tabular}{l|rrc}
  \hline
  Domain & Raw & $\operatorname{sim}>0.6$ & $\operatorname{sim}\in[0.4, 0.6]$ \\
  \hline
  ECB & \num{195956} &   51.73\% & 14.06\% \\
  EMEA & \num{373235} &  65.68\% & 12.65\% \\
  Epps & \num{2009489} & 10.12\% & 25.30\% \\
  GNOME & \num{55391} &  39.31\% & 11.06\% \\
  JRC & \num{503437} &   50.87\% & 16.67\% \\
  KDE & \num{180254} &   36.00\% & 10.81\% \\
  News & \num{151423} &   2.12\% & 9.65\% \\
  PHP & \num{16020} &    34.93\% & 12.38\% \\
  TED & \num{159248} &   11.90\% & 26.64\% \\
  Ubuntu & \num{9314} &  20.32\% & 8.26\% \\
  Wiki & \num{803704} &  19.87\% & 17.32\% \\
  \hline
  Total & \num{4457471} & 24.27\% & 20.00\% \\
  \hline
  \end{tabular}
  }
  \caption{Dataset statistics, with ratios of sentences with at least one TM match for various similarity ranges.\label{tab:data-stat}}
\end{table}

\section{Detailed Results on Each Domain\label{sec:res-detail}}

BLEU and COMET scores for each domain are in Tables~\ref{tab:no-fm-0.6}, \ref{tab:fm-0.6}, \ref{tab:no-fm-0.4}, \ref{tab:fm-0.4}. The variation in scores across domains is large, confirming that TM matches can be very beneficial for some technical domains (e.g.\ ECB, EMEA, GNOME, KDE, JRC), for which we often find good matches that help to greatly increase the performance. On the other hand, News, Wiki and TED yield less matches, and these only help for both the AR approach and \ourmodel{} when the similarity is high ($\operatorname{sim} > 0.6$).

\begin{table*}[!ht]
  \center
  \scalebox{0.75}{
  \begin{tabular}{l|ccccccccccc|c}
  \hline
  BLEU & ECB & EMEA & Epps & GNOME & JRC & KDE & News & PHP & TED & Ubuntu & Wiki & All \\
  \hline
  copy & 59.8 & 64.5 & 34.4 & 70.3 & 67.6 & 55.3 & 12.0 & 38.6 & 30.8 & 51.6 & 47.4 & 52.6 \\
  \hline
  AR & 58.7 & 53.8 & 55.8 & 55.0 & 68.8 & 53.9 & 27.1 & 18.2 & 62.0 & 54.0 & 65.0 & 51.2 \\
  LevT & 46.6 & 30.7 & 51.8 & 51.0 & 62.3 & 47.0 & 23.6 & 12.5 & 58.7 & 50.0 & 61.9 & 46.5 \\
  \ourmodel & 53.0 & 49.7 & 53.2 & 51.5 & 64.7 & 50.8 & 24.5 & 37.1 & 59.5 & 50.4 & 64.0 & 52.6 \\
  \hline
  COMET & ECB & EMEA & Epps & GNOME & JRC & KDE & News & PHP & TED & Ubuntu & Wiki & All \\
  \hline
  copy & 0.4006 & 0.4625 & -0.0797 & 0.4893 & 0.6893 & 0.1150 & -0.6083 & -0.1977 & -0.4184 & 0.3296 & 0.2843 & 0.1330 \\
  \hline
  AR & 0.6333 & 0.6402 & 0.8137 & 0.7190 & 0.9057 & 0.5116 & 0.3241 & -0.0556 & 0.7848 & 0.7031 & 0.7786 & 0.6143 \\
  LevT & 0.4251 & 0.1322 & 0.7460 & 0.6181 & 0.8291 & 0.3879 & 0.2037 & -0.6139 & 0.6912 & 0.5636 & 0.6947 & 0.4251 \\
  \ourmodel & 0.5637 & 0.5559 & 0.7513 & 0.6355 & 0.8477 & 0.4218 & 0.1660 & -0.0980 & 0.6929 & 0.5768 & 0.7335 & 0.5314 \\
  \hline
  \end{tabular}
  }
  \caption{BLEU and COMET scores for each domain, the task is \textbf{standard MT} with $\mathbf{\operatorname{sim}>0.6}$. \textit{All} is computed by concatenating all test sets ($11$k sentences in total). \textit{Copy} refers to copying the TM match into the output.\label{tab:no-fm-0.6}}
\end{table*}

\begin{table*}[!ht]
  \center
  \scalebox{0.75}{
  \begin{tabular}{l|ccccccccccc|c}
  \hline
  BLEU & ECB & EMEA & Epps & GNOME & JRC & KDE & News & PHP & TED & Ubuntu & Wiki & All \\
  \hline
  copy & 59.8 & 64.5 & 34.4 & 70.3 & 67.6 & 55.3 & 12.0 & 38.6 & 30.8 & 51.6 & 47.4 & 52.6 \\
  \hline
  AR & 71.9 & 72.0 & 58.9 & 80.1 & 83.2 & 67.3 & 28.8 & 44.7 & 63.3 & 67.6 & 68.6 & 67.1 \\
  LevT & 62.4 & 53.8 & 55.5 & 77.5 & 78.8 & 63.3 & 26.1 & 28.7 & 60.2 & 66.0 & 67.1 & 60.4 \\
  ~~+tgt TM & 60.2 & 63.8 & 34.8 & 69.6 & 67.7 & 54.5 & 12.5 & 38.8 & 31.1 & 52.1 & 47.5 & 52.8 \\
  \ourmodel & 69.8 & 72.2 & 56.0 & 78.1 & 82.2 & 68.2 & 26.0 & 44.1 & 60.3 & 66.3 & 68.7 & 65.9 \\
  \hline
  COMET & ECB & EMEA & Epps & GNOME & JRC & KDE & News & PHP & TED & Ubuntu & Wiki & All \\
  \hline
  copy & 0.4006 & 0.4625 & -0.0797 & 0.4893 & 0.6893 & 0.1150 & -0.6083 & -0.1977 & -0.4184 & 0.3296 & 0.2843 & 0.1330 \\
  \hline
  AR & 0.7288 & 0.7211 & 0.8223 & 0.9143 & 0.9954 & 0.6299 & 0.3318 & 0.0801 & 0.7910 & 0.8610 & 0.8110 & 0.6985 \\
  LevT & 0.5647 & 0.3384 & 0.7608 & 0.8617 & 0.9355 & 0.5683 & 0.2443 & -0.2183 & 0.7203 & 0.8091 & 0.7618 & 0.5767 \\
  ~~+tgt TM & 0.4086 & 0.4573 & -0.0075 & 0.5230 & 0.7062 & 0.1437 & -0.5679 & -0.1957 & -0.3328 & 0.3710 & 0.3008 & 0.1639 \\
  \ourmodel & 0.6792 & 0.7003 & 0.7591 & 0.8699 & 0.9696 & 0.6106 & 0.2093 & 0.0353 & 0.6923 & 0.8155 & 0.7614 & 0.6454 \\
  \hline
  \end{tabular}
  }
  \caption{BLEU and COMET scores for each domain, the task is \textbf{MT with TMs} with $\mathbf{\operatorname{sim}>0.6}$. \textit{All} is computed by concatenating all test sets ($11$k sentences in total). \textit{Copy} refers to copying the TM match into the output.\label{tab:fm-0.6}}
\end{table*}

\begin{table*}[!ht]
  \center
  \scalebox{0.75}{
  \begin{tabular}{l|ccccccccccc|c}
  \hline
  BLEU & ECB & EMEA & Epps & GNOME & JRC & KDE & News & PHP & TED & Ubuntu & Wiki & All \\
  \hline
  copy & 47.3 & 47.6 & 12.7 & 52.6 & 53.0 & 42.7 & 5.8 & 29.7 & 8.2 & 35.1 & 13.0 & 34.5 \\
  \hline
  AR & 52.3 & 52.7 & 44.7 & 54.4 & 64.7 & 53.2 & 30.0 & 17.9 & 41.7 & 49.2 & 42.2 & 46.1 \\
  LevT & 40.7 & 31.4 & 42.6 & 51.0 & 59.8 & 46.8 & 27.6 & 11.9 & 38.7 & 45.7 & 40.2 & 40.8 \\
  \ourmodel & 47.9 & 47.7 & 41.5 & 51.6 & 61.1 & 50.1 & 26.8 & 34.3 & 38.0 & 46.8 & 41.0 & 45.7 \\
  \hline
  COMET & ECB & EMEA & Epps & GNOME & JRC & KDE & News & PHP & TED & Ubuntu & Wiki & All \\
  \hline
  copy & 0.0310 & 0.1527 & -0.7608 & 0.1416 & 0.1919 & -0.1703 & -0.9719 & -0.6279 & -1.1419 & -0.1837 & -0.8222 & -0.3784 \\
  \hline
  AR & 0.5229 & 0.5920 & 0.7735 & 0.7048 & 0.8834 & 0.5522 & 0.4688 & -0.1819 & 0.5501 & 0.6363 & 0.4157 & 0.5379 \\
  LevT & 0.2908 & 0.1245 & 0.6996 & 0.5956 & 0.8069 & 0.4140 & 0.3567 & -0.7332 & 0.3979 & 0.5194 & 0.3011 & 0.3429 \\
  \ourmodel & 0.4370 & 0.5231 & 0.6515 & 0.6205 & 0.8116 & 0.4576 & 0.2948 & -0.2343 & 0.3655 & 0.5035 & 0.2600 & 0.4263 \\
  \hline
  \end{tabular}
}
  \caption{BLEU and COMET scores for each domain, the task is \textbf{standard MT} with $\mathbf{\operatorname{sim}\in[0.4, 0.6]}$. \textit{All} is computed by concatenating all test sets ($11$k sentences). \textit{Copy} refers to copying the TM match into the output.\label{tab:no-fm-0.4}}
\end{table*}

\begin{table*}[!ht]
  \center
  \scalebox{0.75}{
  \begin{tabular}{l|ccccccccccc|c}
  \hline
  BLEU & ECB & EMEA & Epps & GNOME & JRC & KDE & News & PHP & TED & Ubuntu & Wiki & All \\
  \hline
  copy & 47.3 & 47.6 & 12.7 & 52.6 & 53.0 & 42.7 & 5.8 & 29.7 & 8.2 & 35.1 & 13.0 & 34.5 \\
  \hline
  AR & 62.3 & 62.8 & 44.9 & 69.6 & 75.4 & 62.1 & 29.9 & 39.2 & 42.6 & 58.1 & 43.9 & 55.7 \\
  LevT & 52.3 & 47.1 & 42.7 & 65.7 & 71.9 & 57.6 & 27.5 & 23.8 & 39.0 & 55.0 & 40.8 & 49.3 \\
  ~~+tgt TM & 47.4 & 48.0 & 13.2 & 53.2 & 53.5 & 42.9 & 6.0 & 29.7 & 9.1 & 37.1 & 13.2 & 35.0 \\
  \ourmodel & 59.7 & 61.9 & 41.4 & 68.1 & 73.0 & 61.4 & 26.4 & 39.1 & 37.5 & 56.1 & 39.7 & 53.3 \\
  \hline
  COMET & ECB & EMEA & Epps & GNOME & JRC & KDE & News & PHP & TED & Ubuntu & Wiki & All \\
  \hline
  copy & 0.0310 & 0.1527 & -0.7608 & 0.1416 & 0.1919 & -0.1703 & -0.9719 & -0.6279 & -1.1419 & -0.1837 & -0.8222 & -0.3784 \\
  \hline
  AR & 0.5814 & 0.6607 & 0.7740 & 0.8380 & 0.9220 & 0.6217 & 0.4741 & -0.1140 & 0.5543 & 0.7453 & 0.4344 & 0.5900 \\
  LevT & 0.4283 & 0.2846 & 0.6998 & 0.7697 & 0.8746 & 0.5437 & 0.3660 & -0.4900 & 0.4107 & 0.6676 & 0.2910 & 0.4404 \\
  ~~+tgt TM & 0.0487 & 0.1569 & -0.7208 & 0.1883 & 0.2167 & -0.1151 & -0.9508 & -0.6205 & -1.0949 & -0.1234 & -0.8100 & -0.3478 \\
  \ourmodel & 0.5102 & 0.6281 & 0.6368 & 0.8142 & 0.8741 & 0.5814 & 0.2781 & -0.1853 & 0.3523 & 0.6727 & 0.2172 & 0.4889 \\
  \hline
  \end{tabular}
  }
  \caption{BLEU and COMET scores for each domain, the task is \textbf{MT with TMs} with $\mathbf{\operatorname{sim}\in[0.4, 0.6]}$. \textit{All} is computed by concatenating all test sets ($11$k sentences). \textit{Copy} refers to copying the TM match into the output.\label{tab:fm-0.4}}
\end{table*}

\end{document}